\documentclass[letterpaper, 10 pt, conference]{ieeeconf}  

\IEEEoverridecommandlockouts                              

\usepackage{graphics} 

\usepackage{amsmath}
\usepackage{pdfpages}
\usepackage{booktabs}       
\usepackage{graphicx}
\usepackage{amssymb}
\usepackage{hyperref}

\title{\LARGE \bf
MMPI: a Flexible Radiance Field Representation by Multiple Multi-plane Images Blending
}

\newcommand{\ourmethod}{MMPI}

\author{Yuze He$^{1}$, Peng Wang$^{2}$, Yubin Hu$^{1}$, Wang Zhao$^{1}$, Ran Yi$^{3}$, Yong-Jin Liu$^{1}$,~\IEEEmembership{Senior Member,~IEEE}\\and Wenping Wang$^{4}$,~\IEEEmembership{Fellow,~IEEE}
\thanks{$^{1}$Y. He, Y. Hu, W. Zhao, and Y.-J. Liu are with the Department
of Computer Science and Technology, Tsinghua University, China
        {\tt\small \{hyz22, huyb20, zhao-w19\}@mails.tsinghua.edu.cn, liuyongjin@tsinghua.edu.cn }}%
\thanks{$^{2}$P. Wang is with the Department of Computer Science, The University of Hong Kong, Hong Kong
        {\tt\small totoro97@outlook.com}}%
\thanks{$^{3}$R. Yi is with the Department of Computer Science and Engineering, Shanghai Jiao Tong University, China
        {\tt\small ranyi@sjtu.edu.cn}}
\thanks{$^{4}$W. Wang is with the Department of Computer Science and Engineering, Texas A\&M University, USA
        {\tt\small wenping@tamu.edu}}
}

\begin{document}

\maketitle
\thispagestyle{empty}
\pagestyle{empty}

\begin{abstract}

This paper presents a flexible representation of neural radiance fields based on multi-plane images (MPI), for high-quality view synthesis of complex scenes.
MPI with Normalized Device Coordinate (NDC) parameterization is widely used in NeRF learning for its simple definition, easy calculation, and powerful ability to represent unbounded scenes.
However, existing NeRF works that adopt MPI representation for novel view synthesis can only handle simple forward-facing unbounded scenes (e.g., the scenes in the LLFF dataset), where the input cameras are all observing in similar directions with small relative translations.
Hence, extending these MPI-based methods to more complex scenes like large-range or even 360-degree scenes is very challenging.
In this paper, we explore the potential of MPI and show that MPI can synthesize high-quality novel views of complex scenes with diverse camera distributions and view directions, which are not only limited to simple forward-facing scenes.
Our key idea is to encode the neural radiance field with multiple MPIs facing different directions and blend them with an adaptive blending operation.
For each region of the scene, the blending operation gives larger blending weights to those advantaged MPIs with stronger local representation abilities while giving lower weights to those with weaker representation abilities.
Such blending operation automatically modulates the multiple MPIs to appropriately represent the diverse local density and color information.
Experiments on the KITTI dataset and ScanNet dataset demonstrate that our proposed MMPI synthesizes high-quality images from diverse camera pose distributions and is fast to train, outperforming the previous fast-training NeRF methods for novel view synthesis.
Moreover, we show that MMPI can encode extremely long trajectories and produce novel view renderings, demonstrating its potential in applications like autonomous driving. Our demo video is available at \href{https://youtube.com/watch?v=mbNKwN5urC8}{https://youtube.com/watch?v=mbNKwN5urC8}.

\end{abstract}

\section{INTRODUCTION}

Novel view synthesis (NVS) is a long-standing research problem and has been continuously studied over the past decades, with numerous practical applications such as autonomous driving, virtual reality, etc. 
For the past years, neural radiance field (NeRF)~\cite{mildenhall2020nerf,barron2021mip,tancik2023nerfstudio} has proven to be a powerful tool to model the 3D scenes for the task of novel view synthesis.
Typically, a NeRF model represents a 3D scene by volume densities and view-dependent emissive colors, which is trained by differentiable volume rendering and supervised by the pixel colors of the input posed images. Once training is done, NeRF enables synthesizing photorealistic images from arbitrary viewpoints.

Training a neural radiance field usually requires pre-defining a bound where the scene is represented.
For the rendering of unbounded scenes, a popular strategy is to re-parameterize the unbounded world space to a bounded space, e.g., \cite{mildenhall2020nerf,zhang2020nerf++,barron2022mip}. Among these methods, a widely used technique is Normalized Device Coordinate (NDC) re-parameterization with the multi-plane images (MPI) data arrangement
\cite{wizadwongsa2021nex,sun2022improved}.
The NDC re-parameterization maps an infinitely far view frustum to a unit cube and relocates NeRF's ability to make it consistent with the perspective cameras~\cite{barron2022mip,mildenhall2020nerf}.
MPI benefits from the provided additional depth constraint, and converges more easily than those data structures without a fixed depth, which allows a larger depth range to be modeled.
So far, despite of its powerful representation ability, MPI with NDC mapping is only suitable for simple forward-facing scenes, which assumes all the cameras look along with a similar view direction, and the translations between different cameras are small.
The reason for this limitation is that NDC needs to predefine a view frustum for calculating the mapping, and if a new camera frustum's orientation or translation is far off the predefined one, the mapping is unreliable, leading to severe degradation of the synthesized image quality.

In this paper we address the problem: is it possible to extend MPI to render more complex scenes, e.g., large-range scenes or even 360-degree scenes?
We explore the potential of MPI and present a novel solution to this problem.
We propose Multiple Multi-plane Images (\ourmethod), a flexible representation of neural radiance fields to synthesize high-quality images of complex scenes.
As the name indicates, we encode the scene with a set of multiple MPIs.
By properly organizing and arranging the positions and orientations of those MPIs, our representation is able to cover a wide, unbounded range of the scene of interest.
Based on this representation, we propose a reliability field for each MPI to blend the sampled colors and densities for a proper volume rendering.
We further propose a two-stage reliability learning scheme to effectively train the MPI and its reliability field. At the first stage each MPI is trained individually, after that all the MPIs are then jointly trained using an adaptive blending technique to learn their reliability at each spatial position.
The adaptive blending gives larger weights for advantaged MPIs with better local representation abilities and lower weights for MPIs with less representation abilities.
In this way, the MPIs collaborate with each other and further increase the rendering quality.

We have conducted extensive experiments on the KITTI dataset and ScanNet dataset. Experimental results demonstrate that MMPI is fast to train (about 40 minutes to converge on a large range scene on a single Nvidia 3090 GPU), and achieves state-of-the-art novel view synthesis quality among fast-training NeRF methods.

\section{RELATED WORKS}
\noindent
{\bf Fast Neural Radiance Fields.} The original NeRF model~\cite{mildenhall2020nerf} is encoded by a multi-layer perceptron (MLP), which usually takes hours or days to converge due to the complex optimization of the deep model and is slow to render an image.
Some works focused on accelerating the rendering speed of NeRFs and achieved real-time rendering by baking or distilling from a pre-trained NeRF model~\cite{yu2021plenoxels,hedman2021baking,reiser2021kilonerf} but such training and baking processes are still slow.
Instead of training a deep neural network, recent works showed that the training process can be greatly accelerated by direct optimization of voxels~\cite{sun2021direct,yu2021plenoxels}, neural hash grids~\cite{muller2022instant,tancik2023nerfstudio,li2022nerfacc} or tensor decomposition~\cite{chen2022tensorf,tang2022compressible}.
In this paper, following DVGO~\cite{sun2021direct} we use MPI with voxel representation to encode the density and colors for speeding up the training process.

\noindent
{\bf NeRFs for unbounded scenes.} Typically, a NeRF model is only able to encode bounded scenes.
To render unbounded scenes, recent works adopt different space parameterization methods to map an unbounded scene to a bounded scene.
A widely used one is Normalized Device Coordinate (NDC) mapping with multi-plane images (MPI). NDC maps an unbounded view frustum to a bounded cube~\cite{mildenhall2020nerf, yu2021plenoxels, chen2022tensorf, sun2021direct}, which can suitably represent a forward-facing scene and is easy to compute.
For unbounded 360-degree scenes, some recent works proposed several reparameterization methods~\cite{barron2022mip, yu2021plenoxels, zhang2020nerf++, neff2021donerf}, which share a similar idea that maps an infinitely large spherical space to a bounded sphere space.
More recently, MERF~\cite{reiser2023merf} proposed a line-preserving mapping for efficient rendering of 360-degree unbounded scenes, and F2-NeRF~\cite{wang2023f} proposed an adaptive space reparameterization method called perspective warping for free trajectories.
MPI is an efficient 3D scene representation format consisting of $L$ image planes located at a set of fixed depths~\cite{zhou2018stereo,SrinivasanTBRNS19,mildenhall2019llff,flynn2019deepview,LiHHHW21,LiHJHHW21}, which has been demonstrated to perform well on forward-facing scenes due to the additional depth constraint, and converges more easily than other alternatives without fixed discretized depths.
In this work, we use NDC mapping for its simple definition and easy calculation, and we propose Multiple MPI blending with an adaptive blending operation to render complex unbounded scenes. 
Nex360~\cite{phongthawee2022nex360} also used multiple MPIs for rendering 360-degree scenes. However, it  
conducts the blending operation on 2D image space and is slow to train, while our method blends the MPIs directly on the 3D space and has a significantly fast convergence speed.

\noindent
{\bf NeRFs for large-scale scenes.}
Since the emergence of NeRF, recent works have tried to extend NeRF to large-scale scenes, which usually requires training a large number of NeRF models and composing those models for large-scale scene representation~\cite{tancik2022block, rematas2022urban, turki2022mega, xiangli2022bungeenerf}.
Our method is orthogonal and compatible with such scene composition methods and can be possibly extended to large-scale urban scenes.

\section{METHOD}

Our goal is to synthesize high-quality images from novel views given several posed images of unbounded scenes as supervision.
We propose Multiple Multi-plane Image Blending (MMPI), a new scene representation based on NeRF \cite{mildenhall2020nerf} for novel view synthesis of complex unbounded scenes. We encode the scene with multiple multi-plane images and design an adaptive blending strategy based on a learnable reliability grid. 

In this section, we first introduce the preliminaries of the method in Sec. \ref{sec:preliminaries}, which includes the basic NeRF pipeline, voxel grid and MPI scene representations. 
In Sec. \ref{sec:blending}, we introduce our Multi-MPI blending operation and provide an efficient method for fused training and rendering of multiple MPIs.
Sec. \ref{sec:reliab} describes how to improve the rendering effect of Multi-MPI Blending by learning the per-voxel reliability.
Moreover, Sec. \ref{sec:centervox} addresses and overcomes the inherent drawbacks of the MPI format by blending with an extra centered voxel grid.

\begin{figure}[th]
  \centering
  \includegraphics[width=1\linewidth]{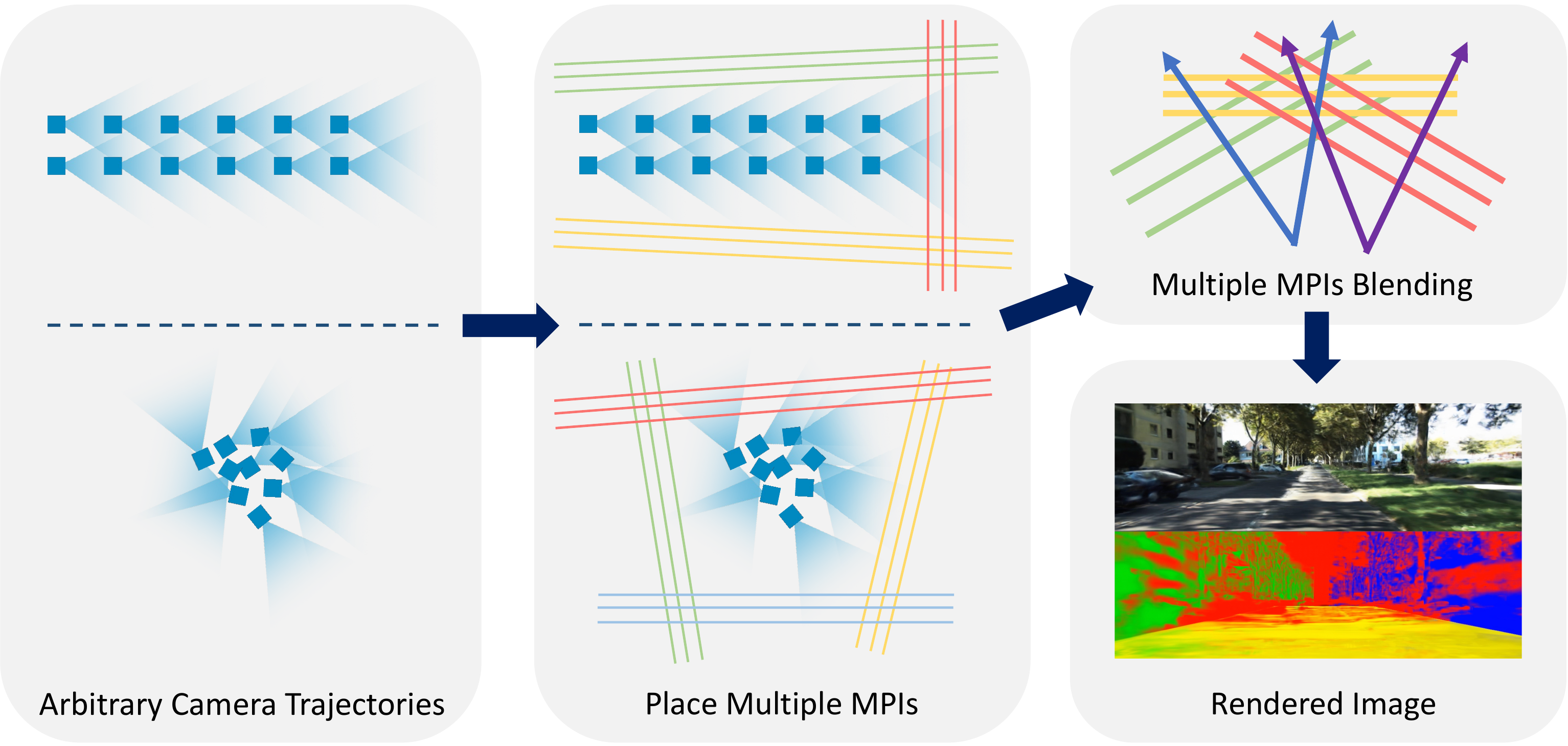}
  \caption{An overview of our proposed MMPI pipeline. We utilize multiple MPIs facing different directions and jointly render them for novel view synthesis. Our MMPI method support blending and rendering any number of arbitrarily located MPI grids to support a wide variety of camera trajectories.}
  \label{fig:pipeline}
\end{figure}

\subsection{Preliminaries} \label{sec:preliminaries}

\noindent
{\bf Neural Radiance Fields.}
Generally, Neural Radiance Fields (NeRF) represent a 3D scene by spatial-variant volume densities with spatial- and view-direction-variant emissive colors, which can be modeled as a learnable function $F_\Theta$ that 
takes the 3-dimensional location of a sampled point $\mathbf{x} = (x; y; z)$ and a 2-dimensional viewing direction $\mathbf{d}=(\theta;\phi)$ as inputs, and outputs density $\sigma$ and color $c$:
\begin{align}
    (\sigma, c) & = F_\Theta(\mathbf{x}, \mathbf{d}).
\end{align}
When rendering a pixel color from a ray of view, a volume-rendering-like formula is employed that involves marching along the ray to determine the color of a pixel $\hat{C}(\mathbf{r})$. 
In the ray marching process, a set of 3D points is sampled along the ray and the synthesized pixel color is integrated by the volume rendering equation from the sampled densities $\sigma_i$ and colors $c_i$ by:
\begin{align} 
\hat{C}(\boldsymbol{r}) &=\sum_{i=1}^N T_i (1-\exp (-\sigma_i \delta_i)) c_i, \nonumber\\
                       &\mbox{where } T_i = \exp \left(-\sum_{j=1}^{i-1}\sigma_j\delta_j\right)
\end{align}
where $\delta_i=t_{i+1}-t_i$ is the distance between adjacent samples.
This rendering process is differentiable, and therefore, the model can be optimized by minimizing the difference between the observed and rendered colors.

\noindent
{\bf Voxel Grid Representation.}
To achieve faster convergence, we follow DVGO~\cite{sun2021direct} and incorporate an explicit voxel grid for modeling 3D information in the scene representation. Different from traditional NeRF architecture, which uses MLP to predict density from 3D coordinates, we construct a learnable 3D density voxel grid $\mathbf{V_\sigma}$ with a voxel number of $N_x\times N_y\times N_z$. We calculate the density $\sigma(\mathbf{x})$ at a particular point $\mathbf{x}$ through trilinear interpolation of 3D coordinates:
\begin{align}
    & \sigma(\mathbf{x})=interp(\mathbf{x}, \mathbf{V_\sigma}), \\
    & interp(\mathbf{x}, \mathbf{V_\sigma}):(R^3, R^{1\times N_x\times N_y\times N_z})\to \mathbb{R},
\end{align}
To acquire color information, we construct a D-channel learnable 3D feature voxel grid $\mathbf{V}_{feat}$ using an explicit-implicit hybrid representation in order to achieve efficient and view-dependent rendering:
\begin{align}
    & \mathcal{F}(\mathbf{x}) =interp(\mathbf{x}, \mathbf{V}_{feat}), \\
    & interp(\mathbf{x}, \mathbf{V}_{feat}) :(R^3, R^{D\times N_x\times N_y\times N_z})\to \mathbb{R}^D,
\end{align}
where $\mathcal{F}(\mathbf{x}) $ contains the color information for different viewing angles. The final color is determined by the feature $\mathcal{F}(\mathbf{x})$, the 3D coordinates $\mathbf{x}$ and the viewing-direction $\mathbf{d}$ together by passing a shallow MLP network $MLP^{(rgb)}_\Theta$:
\begin{align}
    c(\mathbf{x},\mathbf{d}) = MLP^{(rgb)}_\Theta(\mathcal{F}(\mathbf{x}),\mathbf{x}, \mathbf{d}),
\end{align}
where $c$ is the view-dependent color emission. Similar to NeRF, we also use the positional encoding strategy for $\mathbf{x}$ and $\mathbf{d}$ as additional input to MLP.

\noindent
{\bf MPI Representation.}
The voxel grid representation is limited in its ability to accurately represent unbounded 3D scenes, as objects that are located beyond the range of the grids cannot be properly modeled. To address this issue, we adopt the approach proposed by DVGOv2~\cite{sun2022improved}, which reorganizes the voxel grid to the Multi-Plane Images (MPI) format, with a collection of $L$ RGB-density image planes at fixed depths. By linearly sampling disparity from the depth of a specified near plane to $\infty$ in the original space, we place a learnable image plane at each sampling depth. These planes collectively form a MPI, which allows us to address the challenge of representing infinitely far objects. We leverage the same parameterization method as NeRF to warp the forward-facing scene into the normalized device coordinate (NDC) space. Then, we evenly select several image planes within the $z\in[-1,1]$ range in the NDC space to achieve the desired effect above.

The representations for density and color features are essentially similar to the voxel grid mentioned previously. However, for obtaining density and feature by coordinates, interpolation is only carried out along the $x$ and $y$ axes, and not along the $z$-axis
(when sampling points for each ray, only points located at the depth of the image planes are selected, and no other points are included).

\subsection{Multiple MPI Blending}
\label{sec:blending}
Multi-Plane Images (MPI) is an efficient 3D scene representation format consisting of $L$ image planes, each located at a fixed depth $d_i$. Because of the additional depth constraint provided, MPI performs better for forward-facing scenes shot in nearly the same direction, and converges more easily than those data structures without a fixed depth, which allows a larger depth range to be modeled. However, the MPI data structure also has many inherent defects. First, MPI can only cover one direction of the scene; moreover, the depth of the image plane selected by MPI is discrete, but the depth of the scene, in reality, is continuous. When using MPI to model scenes with greater depth variation, no matter how the image planes are arranged, there will be a region with 
excessive
depth gap, and the depth of the scene in this region cannot be correctly predicted.

\begin{figure}[th]
  \centering
  \includegraphics[width=0.8\linewidth]{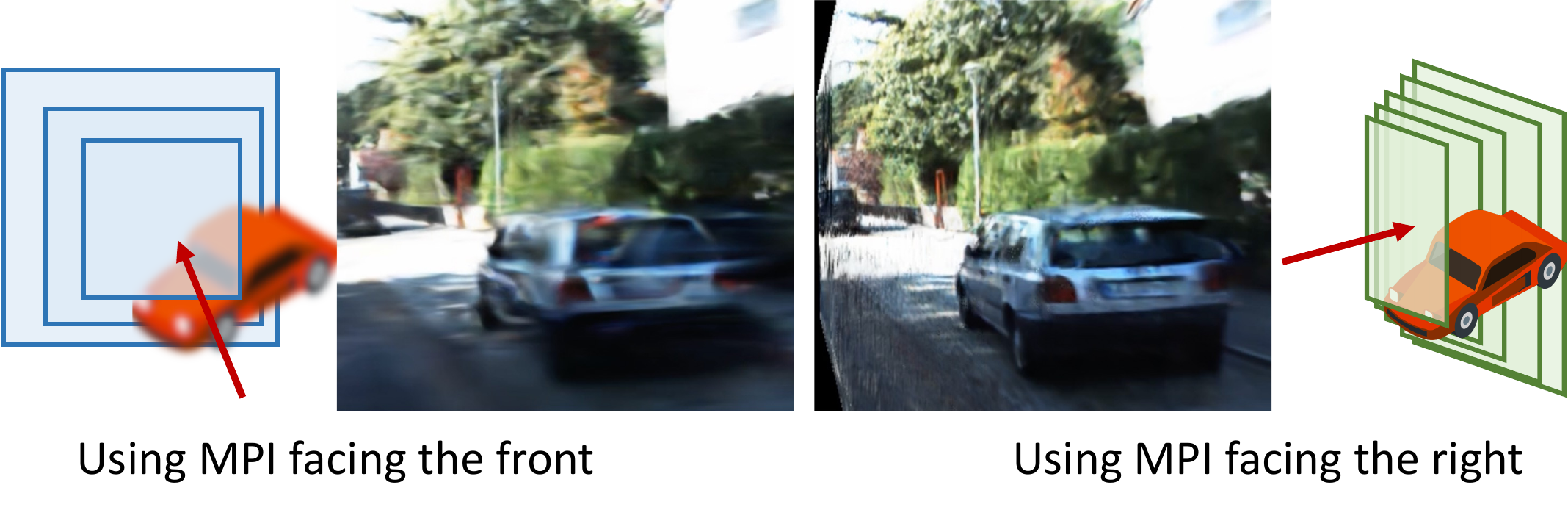}
  \caption{Illustration of the rendering quality when using different-facing MPI grids.}
  \label{fig:diffdir}
\end{figure}

\begin{figure}[th]
  \centering
  \includegraphics[width=0.9\linewidth]{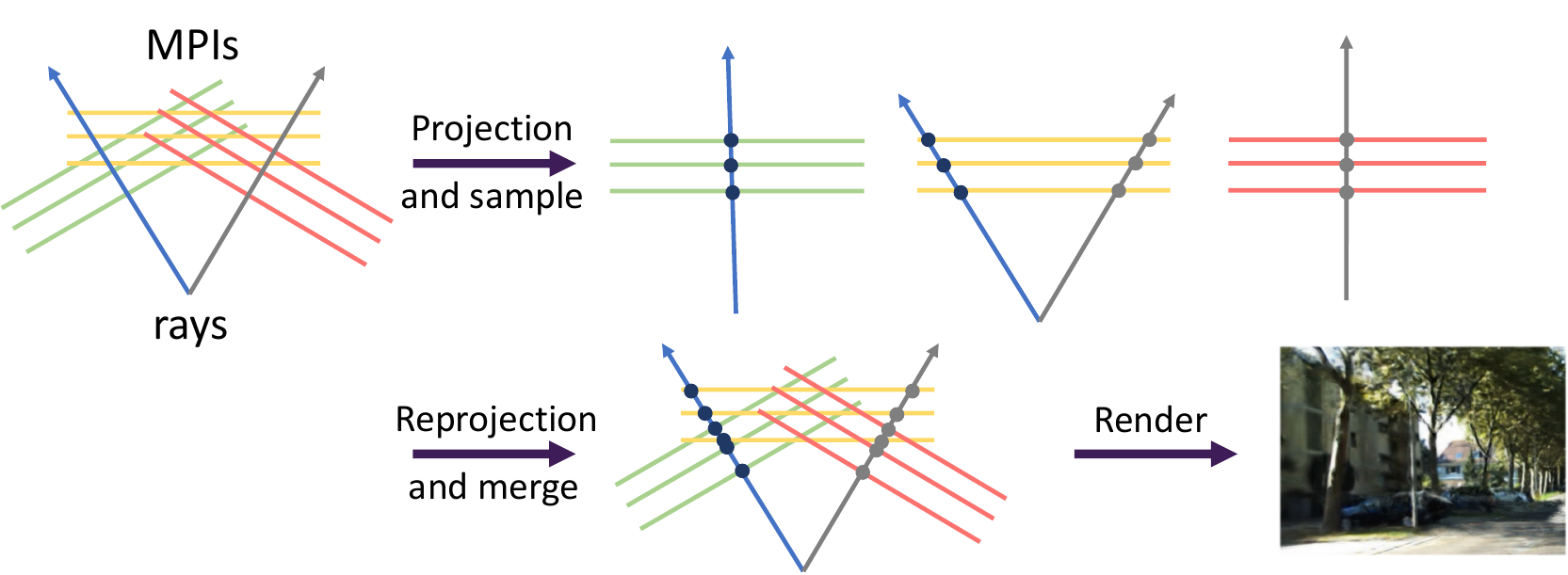}
  \caption{An 2D illustration of our Multiple MPI blending pipeline. We first project camera rays to each MPI's NDC space, sample points and obtain their alpha and color values, and finally all the points are projected into the world coordinate system and merged by their distance to the camera.}
  \label{fig:blending}
\end{figure}

Our MMPI method is based on a key observation that changing the MPI's orientation can result in better modeling of a specific region of the scene. For instance, in the scene of KITTI dataset as shown in Fig. \ref{fig:diffdir}, when we use the MPI facing the front of the scene, we find that the cars on the right side are blurrier, while when we use the MPI facing the right side of the scene, the image quality of the cars on the right side becomes significantly better. This is because the depth change of the cars in the front and back direction is too big, and the cars fall into the sparse area of planes in the MPI facing the front of the scene; while the depth change in the left and right direction is smaller and many objects are closer, the cars fall into the dense area of planes in the MPI facing the right side of the scene. Similarly, small objects such as roadside poles and tree trunks are more easily modeled by MPI facing left or right sides; the surfaces of some buildings need to be predicted with continuously changing depths when using MPI facing the front of the scene, but only a constant depth needs to be predicted by MPI facing left or right sides, thus improving the quality of scene modeling.

Based on the above observation, a reasonable approach to improve the modeling quality of a scene is to leverage {\it multiple MPIs} facing different directions and render them jointly, which allows for the combination of their individual advantages in a cohesive manner. Thus we propose Multiple MPI Blending (MMPI), which supports any number of MPI grids with any orientation for mixed rendering.

We place $K$ MPIs with different orientations in 3D space, each with a grid in its own NDC space, modeling all objects from the near plane to infinity. For a particular training viewpoint, we first generate a set of rays $\{ \mathbf{r}_s \}$ based on all image pixels in that viewpoint. Then, for each ray $\mathbf{r}$ in the set, we need to obtain its intersection points with all MPI planes and arrange them in the order of their distance from the camera. This process can be done using traditional planar intersecstion algorithms (\textit{e.g.}, DDA \cite{marschner2018fundamentals}), but it can significantly slow down the training and rendering process. Instead, we propose the ``sample and merge" strategy (Fig.~\ref{fig:blending}), which effectively improves the sampling speed.

{\bf ``Sample and merge" strategy.} 
For each ray $ \mathbf{r}=(\mathbf{o},\mathbf{d}) $, we first project it into each MPI's NDC space 
and sample the corresponding MPI $\mathcal{M}_i$ to get a set of intersection points $\{ p_j \}_i$. The projected ray direction is then denoted as $Proj_i(\mathbf{d})$. Due to the ambiguity of the z-axis direction in the NDC space transformation, we need to perform an additional forward-facing check to mask out the points with z-component less than zero (which are back to the current MPI $\mathcal{M}_i$) in the $Proj_i(\mathbf{d})$ orientation. Then we input $\{p_j\}_i$ and $Proj_i(\mathbf{d})$ into the density and feature grid of the corresponding MPI to obtain the density and color sets $\{\sigma_j\}_i$, $\{c_j\}_i$, and further calculate the alpha set $\{\alpha _j\}_i$ by:
\begin{align}
    \{\sigma_j\}_i & =interp(\{p_j\}_i, \mathbf{V}_{\sigma,i}), \\
    \{\mathcal{F}_j\}_i & =interp(\{p_j\}_i, \mathbf{V}_{feat,i}), \\
    \{c_j\}_i & = MLP^{(rgb)}_\Theta(\{\mathcal{F}_j\}_i, \{p_j\}_i, Proj_i(\mathbf{d})), \\
    \{\alpha_j\}_i & =\{ 1-\exp(-\sigma_j\delta_j) \}_i.
\end{align}
Then all point sets $\{p_j\}_i$ are projected to the original 3D space and sorted according to the distance from the camera to obtain an ordered set:
\begin{align}
    & \{p_k, \alpha_k\}_{k=1}^{\sum_l^K n(\{p_j\}_l)}, \\
    & \text{where}\ \vert\vert Proj^{-1}_a(p_{k-1})-\mathbf{o}\vert\vert_2 < \vert\vert Proj^{-1}_b(p_{k})-\mathbf{o}\vert\vert_2, \nonumber\\
    & \ p_{k-1}\in\{p_j\}_a,\ p_{k}\in\{p_j\}_b.
\end{align}
Then we render the pixel color by:
\begin{align}
    \mathbf{\hat C}(\mathbf{r})=\sum_{k=1}^{\sum_l^K n(\{p_j\}_l)} T_k\alpha_kc_k,\quad  \text{where}\ T_k=\prod_{j=1}^{k-1}(1-\alpha_j).
\label{eq:rendering}
\end{align}
The technique of Multi-MPI blending can be expanded to encompass unbounded 360-degree scenes like indoor scenes, where the entire 3D space can be modeled by setting up multiple MPIs and making their frustum range cover the entire 360-degree space.

\subsection{Reliability Learning} \label{sec:reliab}

When there is a large overlap among MPIs, training by directly blending them may result in some degree of degradation. One possible explanation for this is that MPIs with a sparse distribution of planes at certain locations tend to learn faster, whereas MPIs with a denser distribution of planes learn at a slower rate but have better modeling ability. When these MPIs are simultaneously trained, those that are learned first may impede the learning of other MPIs, preventing the finer-level learning of certain parts and leading to a decreased ability to model small objects.

To address the issue mentioned above, we propose a solution called per-voxel reliability learning. We learn the relative confidence of multiple MPIs at each location in an end-to-end manner, and utilize an adaptive blending operation to merge MPIs, which gives larger blending weights to those advantaged MPIs with better local representation abilities while giving lower weights to those with less representation abilities.

{\bf Reliability grid and adaptive blending.} We begin by creating a 1-channel reliability grid $\mathbf{V}_{r,i}$ for each MPI $\mathcal{M}_i$, which has the same size and resolution as the 3D density voxel grid in the NDC space where it is situated. 
For the $i$-th MPI $\mathcal{M}_i$, we obtain the reliability $\mathcal{R}_i(p)$ at a point $p$ in its NDC through trilinear interpolation of 3D coordinates:
\begin{align}
    \mathcal{R}_i(p) & =interp(p, \mathbf{V}_{r,i}), \\
interp(p, \mathbf{V}_{r,i}) & :(R^3, R^{1\times N_x\times N_y \times N_z})\to \mathbb{R}.
\end{align}
For the intersection point $p$ sampled at $\mathcal{M}_i$, to obtain its relative confidence among other MPIs, we project it to the NDC coordinate system where another MPI $\mathcal{M}_j$ is located, and obtain the corresponding reliability $\mathcal{R}_j(p) $ in the $j$-th MPI by:
\begin{align}
    & \mathcal{R}_j(p)=interp( Proj_j(Proj^{-1}_i(p)) ,\mathbf{V}_{r,j} ), \nonumber\\
    & \text{where}\ j\not= i
\end{align}
Note that here the point $p$ does not necessarily fall on the image plane of the target MPI $\mathcal{M}_j$ after two reprojections, but since reliability varies continuously, we use trilinear interpolation at this point, interpolating on all $x,y,z$ axes. After that, the relative confidence $P_i(p)$ of point $p$ over all reliabilities is calculated using the softmax function:
\begin{align}
    P_i(p) = \frac{e^{\mathcal{R}_i(p)}}{\sum_{j=1}^K e^{\mathcal{R}_j(p)} }.
\end{align}
Then the rendering formula Eq.(\ref{eq:rendering}) is updated to the following form:
\begin{align}
    \mathbf{\hat C}(\mathbf{r})=\sum_{k=1}^{\sum_l^K n(\{p_j\}_l)} T_kP_k\alpha_kc_k,\nonumber\\
    \mbox{where}\ \ T_k=\prod_{j=1}^{k-1}(1-P_j\alpha_j).
\end{align}
{\bf Two-stage reliability learning scheme.} To learn the reliability grid, we propose a two-stage learning scheme.
Specifically, in the first training stage, each MPI is trained individually without learning the reliability; after this stage, each MPI has received sufficient training and the expressive ability has been fully explored respectively.
Then in the second stage, we jointly train all the MPIs and learn their reliability at each spatial position in an end-to-end manner.

\subsection{Blending MMPI with centered voxel grid} \label{sec:centervox}

The MPI data format has an inherent drawback: it cannot be effectively viewed from the opposite side and is subject to distortion when viewed from an angle 
nearly parallel to image planes.
Consequently, regardless of the number of MPIs used for blending, it becomes challenging to model objects located in the center of the camera trajectory, thereby limiting the scope of application. To address this issue, we propose the extra use of a centered cube voxel grid $\mathcal{C}$ other than the existing MPIs, which represents the scene information in the area surrounding the camera trajectory and nearby regions. This cube grid is created directly in the world coordinate system and interpolates on the $x$, $y$, and $z$ axes, in contrast to the MPI grid, which only interpolates on the $x$ and $y$ axes when obtaining density and color features. 

In the rendering process, for each ray $\mathbf{r}=(\mathbf{o},\mathbf{d}) $, the point set $\{p_j\}_\mathcal{C}$ is obtained by sampling the cube grid $\mathcal{C}$ directly in the world coordinate system, then color and alpha set $\{c_j\}_\mathcal{C}$, $\{\alpha_j\}_\mathcal{C}$ are obtained by a process similar to that of MPI. Together with the other MPI sampled point sets, they are merged and rendered according to the distance from the camera. Note that since the centered cube grid can only represent part of the scene information in one direction, it is not rendered separately, but always in combination with other MPIs.

\subsection{Training loss}
Our training loss is defined as:
\begin{align}
    \mathcal{L} & = \mathcal{L}_{pho} + \lambda_{pt\_rgb}\mathcal{L}_{pt\_rgb} + \lambda_{bg}\mathcal{L}_{bg} \nonumber\\
    & + \lambda_{dist}\mathcal{L}_{dist} + \lambda_{TV}\mathcal{L}_{TV},
\end{align}
where $\mathcal{L}_{pho}$ is the photometric loss, $\mathcal{L}_{bg}$, $\mathcal{L}_{pt\_rgb}$, $\mathcal{L}_{dist}$, $\mathcal{L}_{TV}$ are background entropy loss, per-point color loss, distortion loss, and total variation loss, respectively.

\section{EXPERIMENTS}

\begin{figure}[t]
  \centering
  \includegraphics[width=0.9\linewidth]{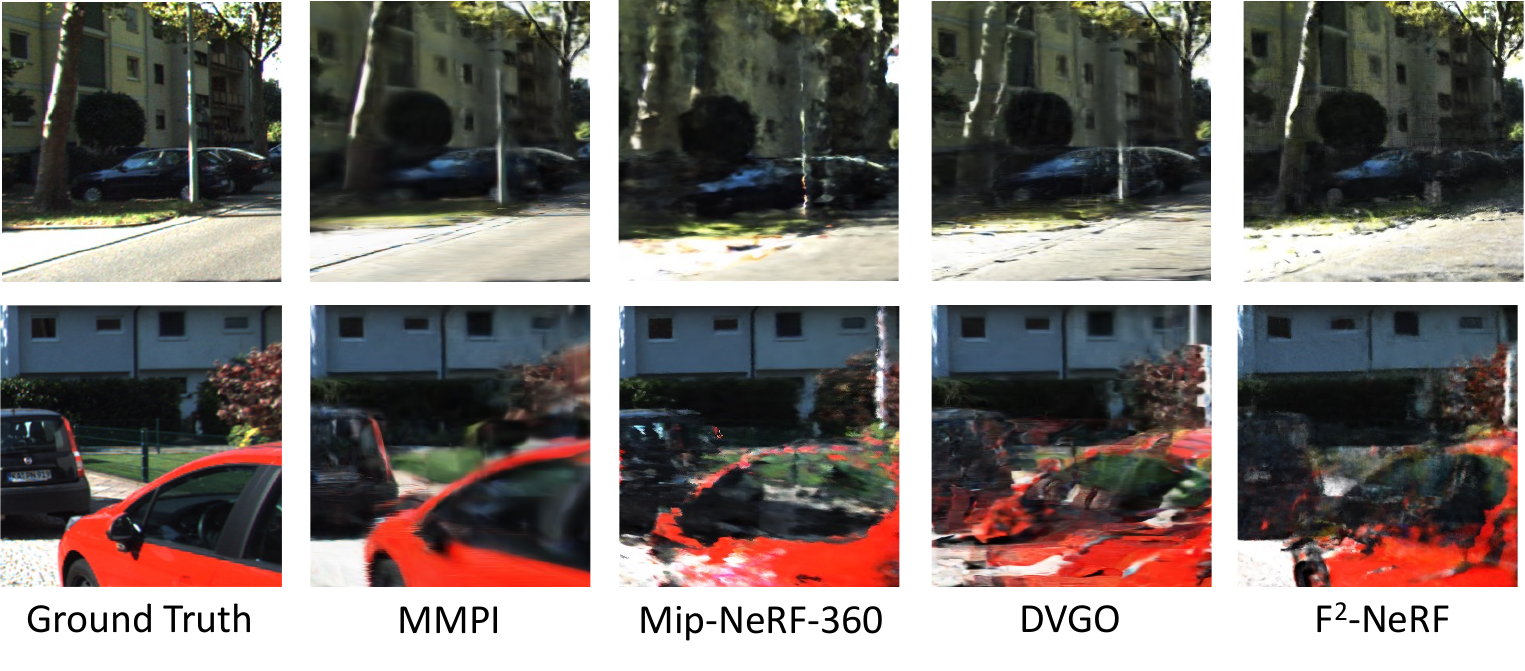}
  \caption{Visual comparison on the KITTI dataset.}
  \label{fig:kitti}
\end{figure}

\begin{figure}[t]
  \centering
  \includegraphics[width=0.9\linewidth]{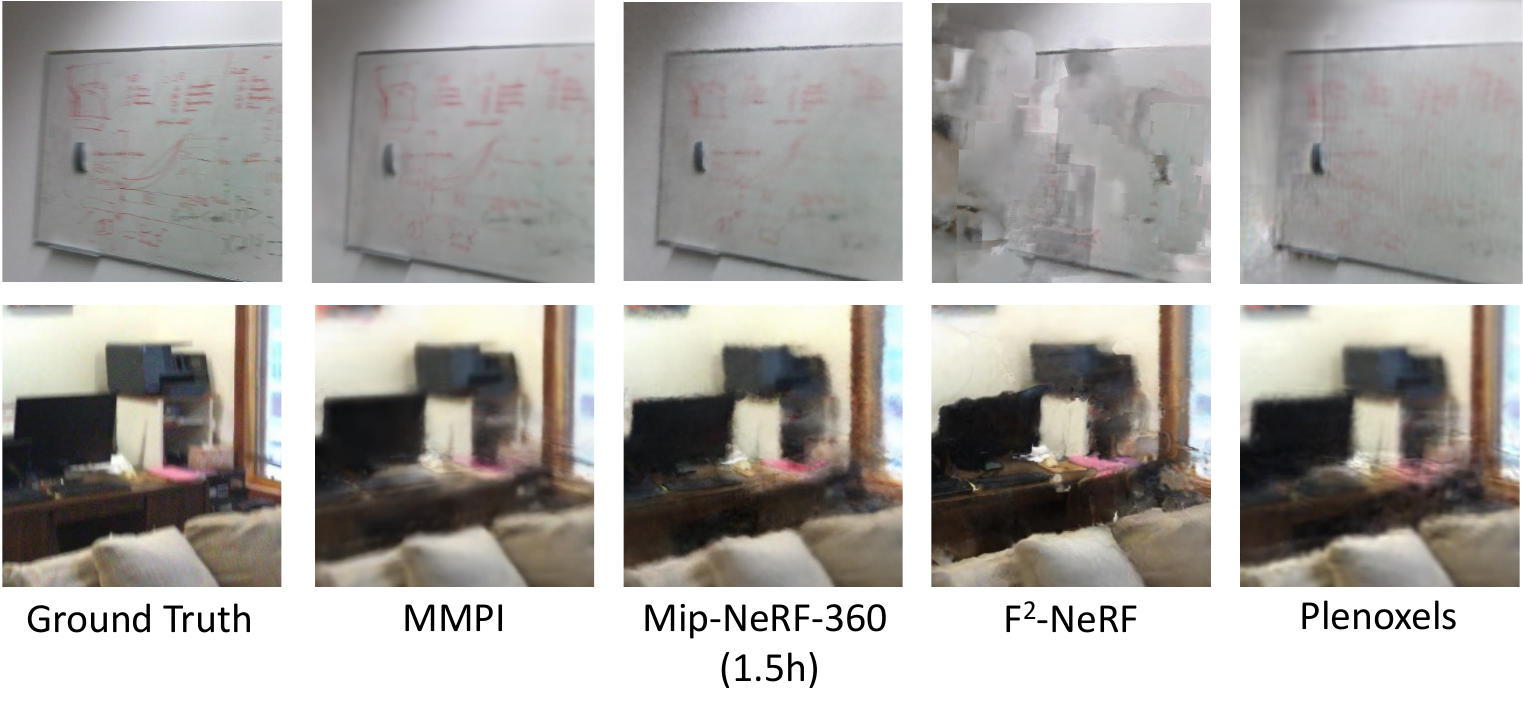}
  \caption{Visual comparison on the ScanNet dataset.}
  \label{fig:scannet}
\end{figure}

\subsection{Datasets and Metrics}

We evaluate our method on two challenging unbounded datasets, KITTI \cite{geiger2012we} and ScanNet \cite{dai2017scannet}. The KITTI dataset contains image sequences of car trajectories captured by the stereo camera facing forward, and we randomly select 5 near-static sequential segments of scenes for evaluation. We use COLMAP \cite{schonberger2016structure} to obtain the ground truth of camera poses, and evenly select 22 images for training, while the remaining 42 images are used for testing. ScanNet is a large dataset containing 1,613 indoor scenes. Following the experimental configurations of NeuRIS \cite{wang2022neuris}, we select eight scenes and evenly choose 1/6 of the images in each scene for training. For testing, we randomly choose 500 images except for the ones used for training. For image quality evaluation, we adopt the three widely used metrics PSNR, SSIM\cite{wang2004image}, and LPIPS$_{VGG}$\cite{zhang2018unreasonable}. We further conduct NVS of extremely long trajectories and provide the details in the supplementary video.

\begin{table}[t]
    \centering
    \caption{Results on the KITTI dataset.}
    \label{tab:kitti}
    \resizebox{\linewidth}{!}{%
    \setlength{\tabcolsep}{2.5pt}
    \begin{tabular}{lccccc}
        \toprule
        Method & Training Time & PSNR$\uparrow$ & SSIM$\uparrow$ & LPIPS$_{VGG}\downarrow$ \\
        \midrule
        NeRF \cite{mildenhall2020nerf}            & 10 hours    & 13.44 & 0.362 & 0.604 \\
        NeRF++ \cite{zhang2020nerf++}          & 22 hours    & 10.82 & 0.297 & 0.681 \\
        mip-NeRF-360 \cite{barron2022mip}    & 11 hours    & \textbf{16.97} & \textbf{0.569} & \textbf{0.448} \\
        \midrule
        Plenoxels \cite{yu2021plenoxels}       & 18 min      & 13.07 & 0.262 & 0.620 \\
        TensoRF \cite{chen2022tensorf}         & 40 min      & 13.33 & 0.292 & 0.608 \\
        F$^2$-NeRF \cite{wang2023f}          & 13 min      & 17.84 & 0.572 & 0.465 \\
        DVGO\cite{sun2022improved}            & 8 min         & 17.58 & 0.558 & 0.495 \\
        MMPI            & 40 min      & \textbf{19.23} & \textbf{0.610} & \textbf{0.464} \\
        \bottomrule
    \end{tabular}
    }
\end{table}

\begin{table}[t]
    \centering
    \caption{Results on the ScanNet dataset.}
    \label{tab:scannet}
    \resizebox{\linewidth}{!}{%
    \setlength{\tabcolsep}{2.0pt}
    \begin{tabular}{lccccc}
        \toprule
        Method & Training Time & PSNR$\uparrow$ & SSIM$\uparrow$ & LPIPS$_{VGG}\downarrow$ \\
        \midrule
        NeRF++ \cite{zhang2020nerf++}          & 28 hours    & 21.78 & 0.717 & 0.556 \\
        mip-NeRF-360 \cite{barron2022mip}    & 36 hours    & \textbf{31.11} & \textbf{0.881} & \textbf{0.277} \\
        \midrule
        Plenoxels \cite{yu2021plenoxels}       & 25 min      & 26.93 & 0.824 & 0.441 \\
        mip-NeRF-360 (short)    & 1.5 hours    & 28.85 & 0.850 & \textbf{0.380} \\
        F$^2$-NeRF \cite{wang2023f}       & 13 min      & 27.67 & 0.835 & 0.387 \\
        MMPI (short)     & 15 min      & 28.19 & 0.836 & 0.445 \\
        MMPI            & 45 min      & \textbf{29.70} & \textbf{0.854} & 0.391  \\
        \bottomrule
    \end{tabular}
    }
\end{table}

\subsection{Implementation Details}

For the KITTI dataset, we use four MPI grids (towards front, left, right and below, respectively) for blending, each MPI has a resolution of 270$\times$270, and a total of 256 depths are selected. For the ScanNet dataset, we use five MPI grids (towards front, back, left, right and below, respectively) and one centered cube grid for blending, where each MPI grid has a resolution of 192$\times$192 and a total of 128 depths selected, and the centered cube grid has a resolution of 160$\times$160$\times$160. When training on the KITTI dataset, we first train each MPI grid individually for a total of 30k iterations, and then freeze all MPIs except the reliability grid for a total of 10k iterations of reliability blending learning. For training on ScanNet, we omit reliability learning due to the small overlap between each MPI and train a total of 30k iterations directly.

\subsection{Comparisons}

We chose several representative works that can synthesize new perspective images of unbounded scenes and conducted quantitative comparisons with our MMPI. The works include NeRF \cite{mildenhall2020nerf}, NeRF++ \cite{zhang2020nerf++}, mip-NeRF-360 \cite{barron2022mip}, Plenoxels \cite{yu2021plenoxels}, TensoRF \cite{chen2022tensorf}, F$^2$-NeRF \cite{wang2023f}, and DVGO \cite{sun2022improved}. The mip-NeRF-360 experiment was conducted on the A100 GPU due to excessive GPU memory, while other experiments and training time statistics were completed on the Nvidia 3090 GPU.

On the KITTI dataset, MMPI outperforms all baseline methods on metrics except LPIPS (Table~\ref{tab:kitti}). As shown in Fig.~\ref{fig:kitti}, the original NeRF and NeRF++ give very vague prediction results, while Plenoxels, DVGO, and TensoRF show structural artifacts, indicating that simple parameterization cannot solve the long-range NVS problem. The perspective warping of F$^2$-NeRF is helpful in solving the NVS of long-range scenes, but there are still many wrong textures. Mip-NeRF-360, due to its own parameterization, predicts some regions too smoothly, loses detailed information, and is also accompanied by artifacts. In contrast, our MMPI can better restore the overall information of the scene while maintaining details such as roadside poles and road shadows.

\begin{table}[t]
    \centering
    \caption{Ablation studies on multiple MPIs and per-voxel reliability learning.}
    \label{tab:ablation}
    \begin{tabular}{lcccc}
        \toprule
        Settings & PSNR$\uparrow$ & SSIM$\uparrow$ & LPIPS$_{VGG}\downarrow$ \\
        \midrule
        Single MPI                      & 16.73 & 0.506 & 0.482 \\
        Multiple MPI w/o reliability    & 17.75 & 0.549 & 0.461 \\
        Multiple MPI with reliability   & \textbf{18.87} & \textbf{0.584} & \textbf{0.458} \\
        \bottomrule
    \end{tabular}
\end{table}

\begin{figure}[t]
  \centering
  \includegraphics[width=0.9\linewidth]{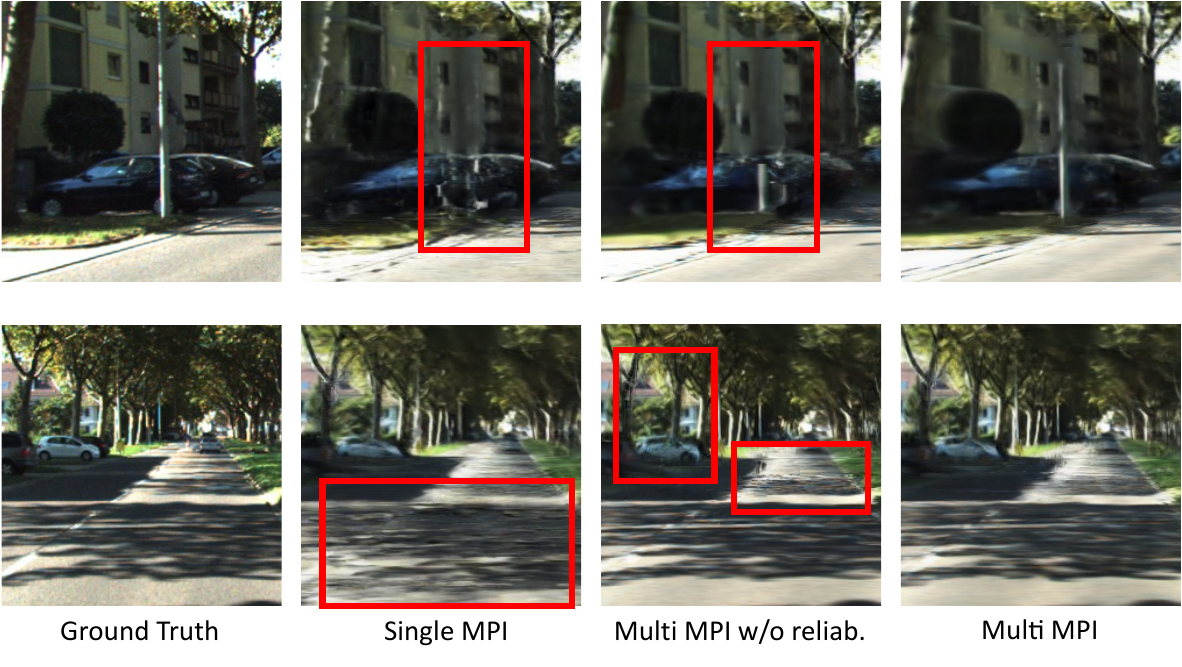}
  \caption{Visual comparison of ablation study.}
  \label{fig:abl}
\end{figure}

We also conducted an evaluation of our method using the challenging 360-degree dataset ScanNet. The results demonstrate that our MMPI approach achieved superior rendering performance in quantitative comparison to all other fast-training NeRF methods (Table~\ref{tab:scannet}). As shown in Fig.~\ref{fig:scannet}, Plenoxels exhibited noticeable banding artifacts, while F$^2$-NeRF presented a large area of false patches. Meanwhile, we observed that mip-NeRF-360 also produced satisfactory results on ScanNet after an extended training period. This is due to its utilization of a larger MLP that can recognize and refine finer textured regions over an extended period of training. However, its prolonged training time poses a limitation on its practical application. 
An early-stopped mip-NeRF-360 with a training time of 1.5 hours will result in a fuzzier outcome.

\subsection{Ablation Studies}

For our ablation studies, we selected a sequence (0096) from the KITTI dataset. We compared the use of a single MPI for training and novel view synthesis (equivalent to using DVGO) with our MMPI approach. To further illustrate the effectiveness of our proposed per-voxel reliability learning for scenes with large overlap among MPIs, we also compared the MMPI method with directly training multiple MPIs without incorporating reliability.

The results in Table~\ref{tab:ablation} indicate that after blending multiple MPIs, the novel view synthesis quality has improved significantly compared to using only a single MPI. Moreover, the introduction of per-voxel reliability learning has further improved image quality. This demonstrates the scene modeling capability of our proposed Multiple MPI blending, as well as the effectiveness of the training pipeline. We provide more qualitative comparisons in Fig. \ref{fig:abl}.

\section{CONCLUSIONS}

We propose Multiple Multi-plane Images (MMPI), a flexible representation of neural radiance fields for novel view synthesis of complex unbounded scenes.
We encode the scene with multiple multi-plane images with different orientations, and design an adaptive blending strategy based on a learnable reliability grid to boost synthesis quality.
By properly organizing and arranging the positions and orientations of those MPIs, our representation is able to cover a wide, unbounded range of the scene of interest.
Extensive experiments on two challenging unbounded datasets demonstrate that our MMPI is fast to train and has superior rendering performance than existing state-of-the-art fast-training NeRF methods.









\bibliographystyle{IEEEtran}
\bibliography{IEEEexample}

\end{document}